%% file: root.tex
\documentclass[letterpaper, 10 pt, conference]{ieeeconf}  

\IEEEoverridecommandlockouts                              

\usepackage{times}
\usepackage{epsfig}
\usepackage{graphicx}
\usepackage{amsmath}
\usepackage{amssymb}
\usepackage{soul}
\usepackage[utf8]{inputenc} 
\usepackage[disable]{todonotes}	
\usepackage{color}	
\usepackage{adjustbox}
\usepackage{multirow}
\usepackage{rotating}
\usepackage{tikz}
\usepackage{xcolor,colortbl}
\definecolor{lgreen}{RGB}{236, 255, 201}
\definecolor{lgray}{RGB}{245,245,245}
\definecolor{lblue}{RGB}{212,241,244}

\usepackage{algorithm}
\usepackage{algpseudocode}  
\usepackage{tabularx}
\usepackage{subcaption}

\usepackage[caption=false,font=footnotesize]{subfig}
\newcommand{\citep}{\cite}

\title{\LARGE \bf
Exploring Camera Encoder Designs for Autonomous Driving Perception}
\author{Barath Lakshmanan, Joshua Chen, Shiyi Lan, Maying Shen, Zhiding Yu, Jose M. Alvarez
}
\begin{document}
\maketitle
\thispagestyle{empty}
\pagestyle{empty}

\begin{abstract}
The cornerstone of autonomous vehicles (AV) is a solid perception system, where camera encoders play a crucial role. Existing works usually leverage pre-trained Convolutional Neural Networks (CNN) or Vision Transformers (ViTs) designed for general vision tasks, such as image classification, segmentation, and 2D detection. Although those well-known architectures have achieved state-of-the-art accuracy in AV-related tasks, e.g., 3D Object Detection, there remains significant potential for improvement in network design due to the nuanced complexities of industrial-level AV dataset. Moreover, existing public AV benchmarks usually contain insufficient data, which might lead to inaccurate evaluation of those architectures.To reveal the AV-specific model insights, we start from a standard general-purpose encoder, ConvNeXt and progressively transform the design. We adjust different design parameters including width and depth of the model, stage compute ratio, attention mechanisms, and input resolution, supported by systematic analysis to each modifications. This customization yields an architecture optimized for AV camera encoder achieving 8.79\% mAP improvement over the baseline. We believe our effort could become a sweet cookbook of image encoders for AV and pave the way to the next-level drive system.
\end{abstract}

\input{intro.tex}

\bibliographystyle{IEEEtranBST/IEEEtran}
\bibliography{bib}

\end{document}

%% file: intro.tex
\section{Introduction}
The ability to accurately understand and react to the surrounding environment is paramount for autonomous driving. This demands a well-designed perception system, with 3D object detection playing a crucial role. Notably, multi-camera-based methods have emerged as the dominant approach for this task, leveraging their affordability, adaptability, and rich 360-degree visual data to pinpoint location, size, and types of surrounding objects such as cars, bikes, and pedestrians. 

\begin{figure}[!t]
  \centering
  \resizebox{\columnwidth}{!}{
  \includegraphics[width=\columnwidth]{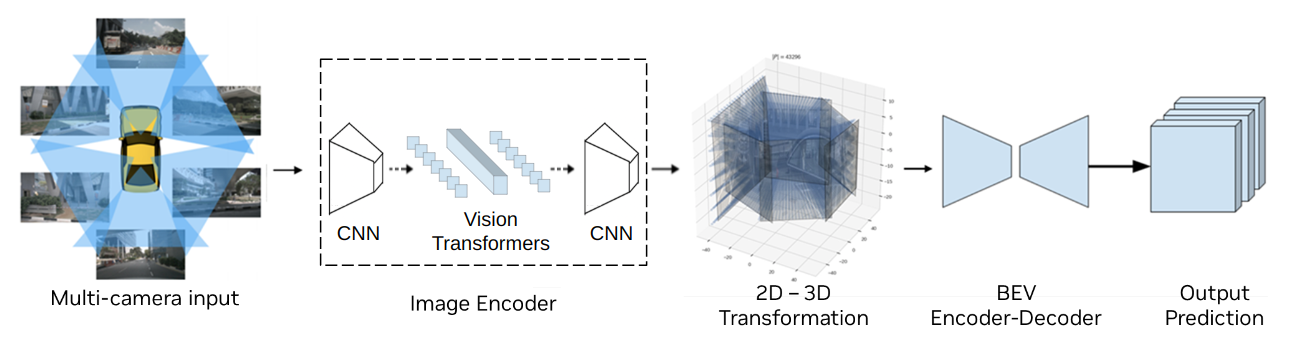}
  }
  \caption{Obstacle 3D Detection Pipeline from Multi-Camera Input. An image encoder extracts relevant features from each input image. The transformation stage projects the 2D image features into a unified 3D space, typically a bird's-eye view (BEV) representation. BEV Encoder-Decoder further processes the 3D features to refine spatial relationships and contextual information. Finally, the prediction stage generates the final 3D obstacle predictions, including their locations, classes, and other relevant attributes.}
  \label{fig:3DObjectDetectionPipeline}
\end{figure}

A strategically deployed multi-camera system, utilizing cameras with diversified field-of-view and resolution, captures overlapping data segments of the vehicle's surrounding environment. Each sensor provides partial environmental information, necessitating data from all the cameras to be collectively processed and transformed. A typical pipeline includes calibration, encoding, 2D-3D transformation, BEV encoder-decoder, and finally, the output prediction as shown in Fig.\ref{fig:3DObjectDetectionPipeline}.  The encoder plays a critical role in this pipeline, employing either a classical CNN model or a modern transformer that also includes CNN models as feature extractor.

While numerous pre-trained models \cite{convnext, coatnet, resnet, vit, liu2021swin} demonstrate impressive performance on general-purpose dataset like Imagenet, leveraging them directly for industrial-level AV dataset has many challenges. First, general-purpose datasets feature many classes, while AV datasets typically involve a restricted set of object classes with substantially more images per class, increasing the class distribution. Second, unlike general-purpose datasets, AV datasets are captured using a rich mix of camera sensor types with contrasting fields of view and resolution. Third, industrial AV datasets have an expansive detection range compared to public datasets like Waymo open dataset \cite{waymo} and nuScenes dataset \cite{nuscenes}, demanding models to have more powerful localization ability, especially for small and distant objects. Lastly, indrustrial-level AV dataset encompass significantly more diverse scenarios compared to public dataset which are limited in scope, hindering accurate evaluation. This fundamental data disparity necessitates domain-specific architecture customization to maximize performance. 

Although there are many network architecture search (NAS) works \cite{wang2022searching, tan2019efficientnet, tan2020efficientdet} for obtaining high-performance and high-efficiency image encoders, we argue that it is important to gather insights into the model design than proposing intricate architecture. Therefore, we start with the state-of-the-art ConvNeXt \cite{convnext}, a simple and popular general-purpose architecture. First, the original ConvNeXt model blocks are modified to facilitate hardware acceleration. Then, leveraging our model insights for AV dataset and conducting systematic analysis to guide the modifications, we alter the design of key components like stages, blocks, channels, stage-compute-ratio, attention mechanism, and input resolution.

Our work shows that customization of the encoder design to the specific characteristics of AV dataset yields 8.79\% relative mAP (mean average precision) improvement over the vanilla model. Further, adapting hybrid architecture brings a separate 1.2\% relative mAP improvement. We also show that the optimized model can effectively scale to create a family of models suitable for different online and offline processing needs.

\begin{figure*}[t!]
  \centering
  \resizebox{0.7\textwidth}{!}{
  \includegraphics[width=0.7\textwidth]{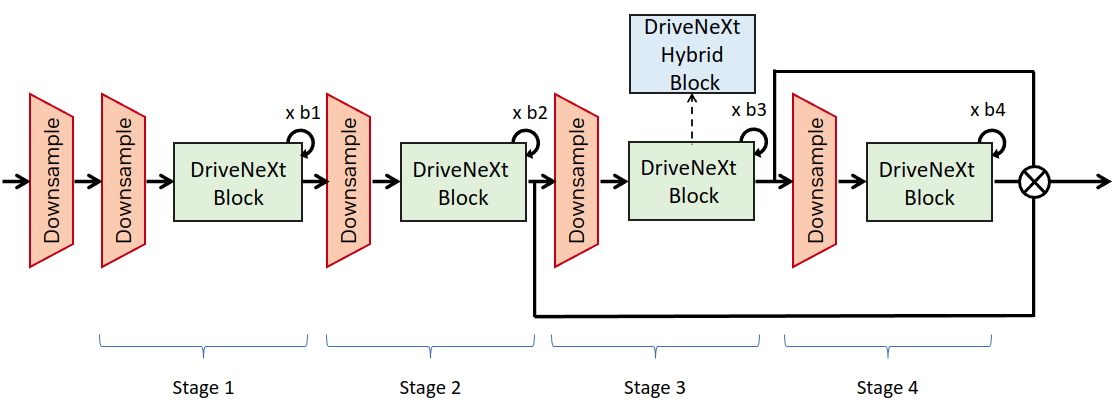}
  }
  \caption{Architecture design of the base model. b1, b2, b3, b4 denote the number of blocks per stage. A hybrid DriveNeXt block replaces the regular block in 3rd stage to realize hybrid architecture. }
  \label{fig:architecture}
\end{figure*}


\section{Related Work}
\label{sect:sota}
Current research on adapting existing deep learning models to specific domains remains limited. While promising for designing optimal architectures, methods like NAS are often constrained by the computational burdens associated with exploring vast search spaces, particularly on large-scale datasets. This section details the state-of-the-art architectures that have demonstrated capabilities in AV and provides a description of publicly available AV datasets.   

\subsection{AI Models for Encoder}
Convolutional neural networks (CNNs) have played a pivotal role in advancing vision models. Early approaches \cite{rcnn, faster_rcnn} relied on multi-stage architecture demonstrating significant performance gains. Single-stage detectors such as SSD \cite{ssd} and YOLO \cite{yolo} perform dense predictions in a single shot, but often at the cost of accuracy. CenterNet \cite{centernet} and FCOS \cite{fcos} introduced a paradigm shift by moving from anchor-based prediction to per-pixel prediction, simplifying the detection pipeline. Deep3DBox \cite{deep3dbox} leveraged 2D detections for 3D bounding box regression via the minimization of 2D-3D projection error. The use of 2D detectors as a starting point for 3D computation has become a standard approach \cite{kehl2017ssd, ku2019monocular}. 

Transformers, with their spatial and temporal attention mechanisms, excel at modeling inter-object relationships across multiple frames. DETR \cite{detr} revolutionized object detection by eschewing traditional convolutional neural networks for a purely transformer-based approach, achieving comparable performance to CNN. Deformable DETR \cite{deformable_detr} builds upon the DETR framework by incorporating deformable attention mechanisms, leading to more accurate bounding box predictions. 

Hybrid architectures \cite{coatnet, moat} aims to bridge the gap between CNNs and Transformers by leveraging both strengths. CNNs excel at extracting local features and capturing spatial relationships within images, while Transformers possess superior capabilities for modeling long-range dependencies and global context. CoAtNet \cite{coatnet}  focuses on optimizing convolutions in shallow layers, demonstrating their effectiveness in capturing local spatial features. MoAT \cite{moat}, introduces Multi-scale Object Attention, a novel attention module that dynamically weights feature maps across multiple scales within the network. This allows MoAT to effectively capture both fine-grained and global context, leading to state-of-the-art performance on image classification tasks. These works highlight the importance of a solid CNN feature extractor and its synergy along with the attention mechanism.

Customizing the architecture of CNN is crucial for optimizing the performance of both pure CNN models and hybrid models. Recent research \cite{convnext} demonstrates that meticulously crafted CNN architectures like ConvNeXt can achieve performance on par with transformers, suggesting that architectural optimization within the CNN paradigm holds significant potential. Furthermore, studies \cite{detnet, rethinking, object} highlight the limitations of repurposing architectures designed for generic image tasks like ImageNet. The field necessitates a shift towards architectures crafted specifically for the nuances of the target domain.


\subsection{Dataset for Autonomous Driving}
Developing autonomous driving algorithms necessitates thorough training and validation across various driving scenarios. This section highlights two prominent datasets, nuScenes and Waymo Open Dataset, with a focus on their camera data characteristics.

\subsubsection{nuScenes}
Captured in Boston and Singapore, nuScenes \cite{nuscenes} boasts diverse urban driving environments with $1000$ scenes, each $20$ seconds long. It includes six synchronized cameras providing $360$° surround view coverage: five monocular cameras with $1600\times900$ resolution and one wide-FOV stereo camera with $1920\times1080$ resolution. The dataset has annotations for $23$ classes, including cars, pedestrians, bicycles, and traffic signs.

\subsubsection{Waymo Open Dataset}
The Waymo Motion Dataset \cite{waymo}, which comprises over $20,000$ labeled driving segments collected from six cities across the United States. Each segment features data from six synchronized cameras mounted on the vehicle's roof, ensuring a $360$° surround view perspective. The cameras have varied resolutions, ranging from $1920\times1080$ to $1280\times720$. The dataset has annotations of $35$ object classes, including vehicles, pedestrians, cyclists, and road elements.

\section{Base architecture design}
\label{sect:method}
In this section, we introduce the architectural design of our baseline camera feature encoder, which serves as the starting point of our exploration. The feature encoder is a key component of any 3D perception system (Fig. \ref{fig:3DObjectDetectionPipeline}) that extracts meaningful features from data using CNNs or Hybrid architectures. Our encoder builds upon ConvNeXt \cite{convnext}, a transformer-like architecture that unlocks impressive performance gains in image classification, object detection, and segmentation among others. Some of the key features from ConvNeXt includes, large initial kernels, inverted bottlenecks, and depthwise separable convolutions. 

The base model is a four-stage hierarchical design for extracting features from the input image, as illustrated in Fig. \ref{fig:architecture}. The encoding process begins with a downsample block that reduces the spatial dimensions of the input image. Each stage consists of a downsample block and a variable number of convolutional blocks. The output from the second and third stages are skip-connected to the final stage. These skip connections help preserve important details from the earlier processing stages, allowing for a more robust and accurate output.

There are a multitude of design parameters in the model, including but not limited to the width and depth of the model, stage compute ratio, attention mechanisms, and input resolution. As detailed in our experimental section (Section \ref{sect:experiments}), these parameters needs to be tuned to the needs of the data as they have a profound influence on the model's ultimate performance.

\section{Experiments and Results}
\label{sect:experiments}
We examine the model's micro-architecture design, fine-tuning its internal components like layers, kernels, normalization, and attention mechanism. Next, we explore macro-architecture design, including number of stages, blocks, block width \& stage-compute-ratio. Then, we focus on altering the input resolution to maximize performance further. Lastly, we experiment with scaling the model to craft a spectrum of architecture with varying complexity for diverse deployment needs.

\subsection{Experimental setup}
To comprehensively evaluate the model efficacy and explore different design choices, we conduct experiments on our internal multi-camera 3D object detection dataset.

\textbf{Dataset:} Our internal large-scale experimental research dataset is annotated with up to four categories: car, truck, person, and bike. While publicly available datasets like Waymo and NuScenes serve well for research purposes, they lack the nuanced data crucial for industrial-level AV like long-range detection using high-resolution cameras and telephoto lenses. Our dataset has a significantly larger volume of data containing 2M scenes and an expansive detection range reaching $250$ meters, surpassing the typical 50-100m range observed in the public dataset. The data is collected using eight cameras strategically placed around the vehicle with a resolution up to $3840\times2160$.

\textbf{Models:} We train the ConvNeXt-derived baseline model and conduct an ablation study, as elaborated later in this section. Additionally, we use two models for reference:
\begin{itemize}
\item A Neural Architecture Search (NAS) based model obtained from \cite{shen2023hardware}, utilizes VGG \cite{vgg} search space but lets the algorithm decide on the number of neurons per layer and the specific number of layers aiming to obtain a good trade-off between accuracy and latency. 
\item YOLO V8 model as a reference off-the-shelf state-of-the-art model with well established architecture.
\end{itemize}

\textbf{Training \&  Evaluation:} To ensure fair comparison and consistency across experiments, all models are trained for $20$ epochs using stochastic gradient descent, with Adam optimizer and learning rate warmup. To evaluate performance, we use mean Average Precision (mAP) across classes, measured on a held-out validation split of the dataset.

\subsection{Micro-architecture Design}

\begin{figure}[t!]
  \centering
  \vspace{0.2cm}
  \resizebox{\columnwidth}{!}{
  \includegraphics[width=1\columnwidth]{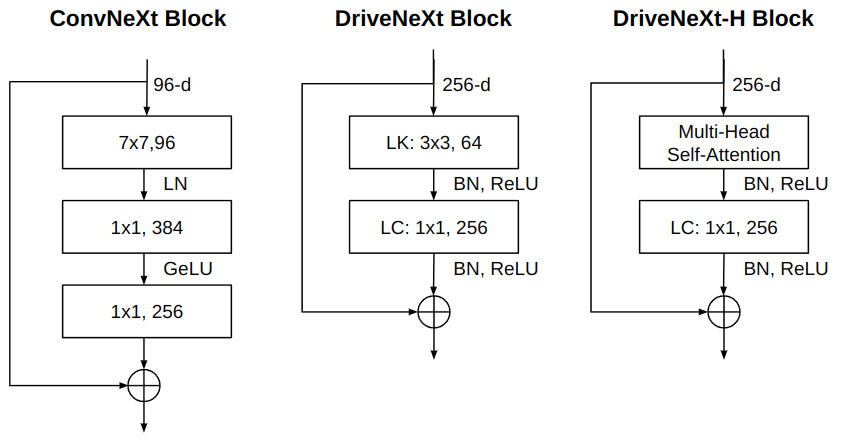}
  }
  \caption{Block evolution from ConvNeXt to DriveNeXt and DriveNeXt-Hybrid.}
  \label{fig:memory}
\end{figure}

\subsubsection{Block design}
\label{sect:blockdesign}
To optimize GPU performance, we modify ConvNeXt blocks with $3\times3$ convolutions for hardware compatibility and replace layer normalization with efficient batch normalization. We adopt a two-layered approach comprising of large kernel (LK) convolution and large channel (LC) convolution with bottleneck design (shrink-then-expand channels) for powerful feature extraction without sacrificing efficiency. This revised architecture, dubbed DriveNeXt, shown in Fig. \ref{fig:memory}, forms the core of our base model.

\begin{figure}[h!]
  \centering
  \resizebox{\columnwidth}{!}{
  \includegraphics[width=\columnwidth]{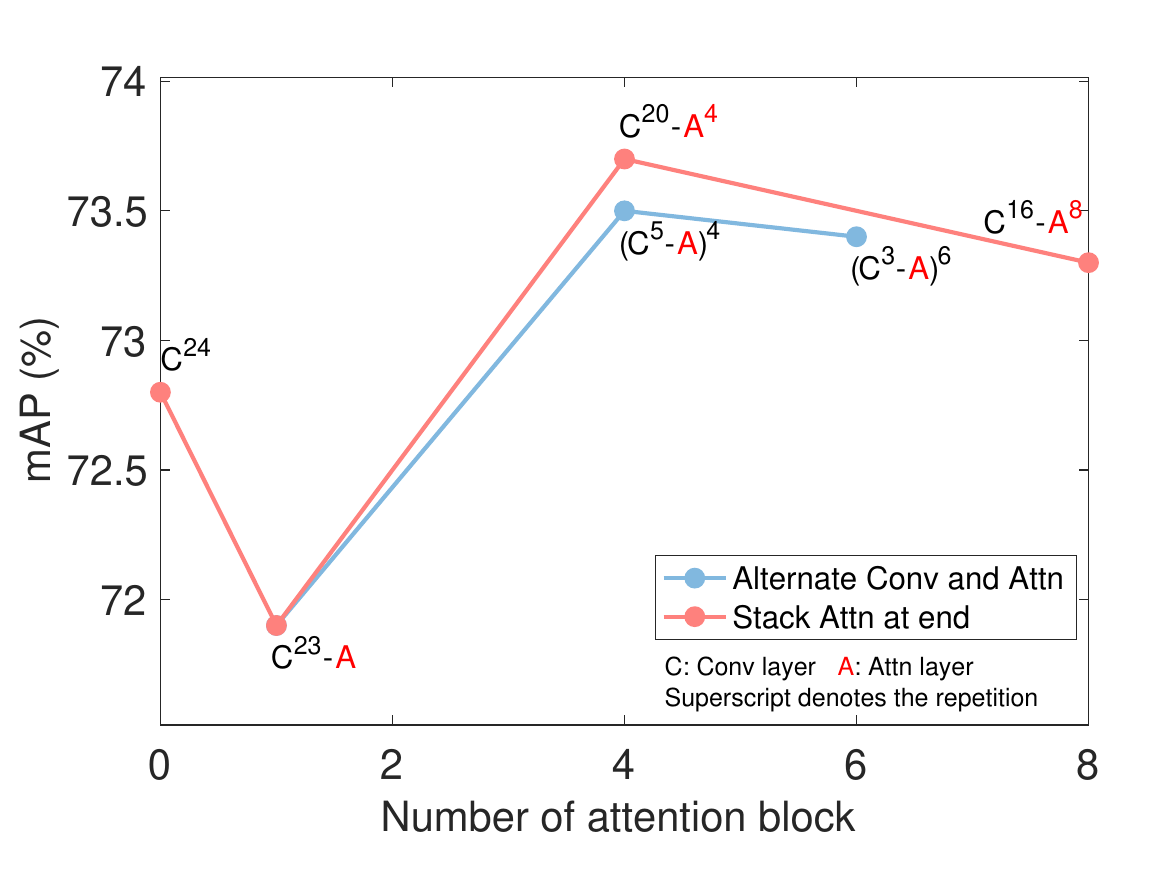}
  }
  \vspace{-0.4cm}
  \caption{Hybrid model ablation: Attention layers and positioning analysis.}
  \label{fig:attention}
\end{figure}

\subsubsection{Adding attention blocks}
Inspired by CoAtNet \cite{coatnet}, we strategically integrated attention modules to enhance feature representation and boost performance. We combined depthwise convolutions with multi-head self-attention in a stacked architecture, see Fig. \ref{fig:memory}. We place attention in third stage as per CoAtNet. Examining the impact of attention depth, four attention blocks emerged as optimal.

Motivated by MOAT, we investigated alternating convolutional and attention layers. However, this offered no substantial gains (see Fig. \ref{fig:attention}). 
Consequently, we adopted a streamlined architecture with attention solely at the end, achieving $0.9\%$ absolute mAP gain. While the hybrid architecture demonstrated a measurable performance boost, to prioritize training efficiency for subsequent explorations, we will use pure convolution based DriveNeXt block design.

\subsection{Macro-architecture Design}
\subsubsection{Changing block width}
Increasing the number of channels in layers is generally expected to provide better accuracy but at increased computation cost. We observe that increasing the number of channels in a large kernel layer beyond a certain level had marginal performance at best. For our model we found $128$ channel count for large kernel layers to be optimal. 

Sequentially increasing channel count across layers has been effective for general-purpose datasets as it demands learning a wider variety of abstract features to accommodate a large number of object classes. With AV dataset we observe that reducing the number of channels in the later stages of the model favorably improves the model speed without affecting mAP.

\subsubsection{Changing number of stages}
we systematically varied the number of stages from $2$ to $5$ while ensuring that the model's complexity remained constant through adjustments in the number of blocks. From our analysis we found that a 4-stage architecture emerges as the optimal choice for our dataset with high resolution and expansive detection range. Architectures with fewer stages exhibit limited hierarchical representation, resulting in a constricted receptive field conducive to detect distant and small objects. Conversely, architectures with a more number of stages have broader receptive field, providing uniform performance across a spectrum of object scales within the image. 

\subsubsection{Changing number of blocks}

\begin{figure}[!t]
  \centering
  \vspace{0.5cm}
  \resizebox{\columnwidth}{!}{
  \includegraphics[width=1.0\linewidth]{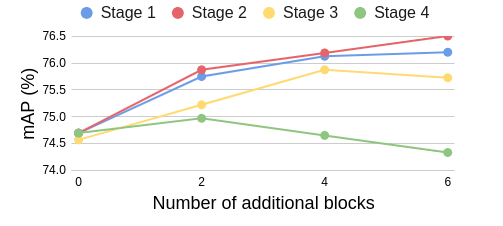}
  }
  \caption{Early CNN stages benefit most from additional blocks, while later stages see diminishing returns.}
  \label{fig:trainlarge}
\end{figure}

We examined the performance impact of increasing the number of blocks in each stage by factors of $2$, $4$, and $6$ in our architecture. The resulting performance trend shown in Fig. \ref{fig:trainlarge}, reveals a critical finding: early stages significantly benefit from additional blocks, while later stages exhibit diminishing returns. 


\subsubsection{Tuning stage compute ratio}
\begin{figure}[!h]
  \centering
  \resizebox{\columnwidth}{!}{
  \includegraphics[width=1.0\linewidth]{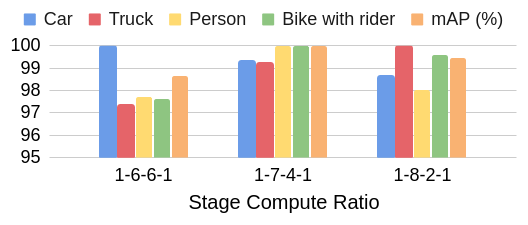}
  }
  \caption{The plot shows the relative performance of models with different stage compute ratios on various classes.}
  \label{fig:perclassresult}
\end{figure}

Optimizing the stage-wise compute ratio, defined as the relative computational expenditure assigned to each stage, is paramount to balancing model performance and efficiency for our specific task: detecting smaller objects in datasets with a limited number of classes. 
Through a systematic exploration of a range of compute ratios, we identified that a $1:7:4:1$ ratio emerged as the most effective configuration for encoder architecture. We show the relative performance of top-performing models across different classes in Fig. \ref{fig:perclassresult}. While no single model dominates across all classes, the stage computes the ratio of $1$-$7$-$4$-$1$ and exhibits balanced performance across all classes.

\subsection{Effect of input resolution}

We explore further performance gains by investigating the model's sensitivity to input resolution. Previous works \cite{resnet, convnext} have successfully employed downsampling techniques to improve computational efficiency during training. As we need our model to be effective in detecting long distance objects, we remove the initial downsample layer to make the network operate at its native resolution i.e., double the input size. The full-resolution model achieved a $2\%$ absolute mAP improvement compared to its downsampled counterpart. This comes at a quadrupled compute cost but improves model performance, particularly when detecting distant objects.

\input{main_result}

\begin{figure}[h!]
  \centering
  \vspace{0.2cm}
  \resizebox{0.8\columnwidth}{!}{
  \includegraphics[width=\columnwidth]{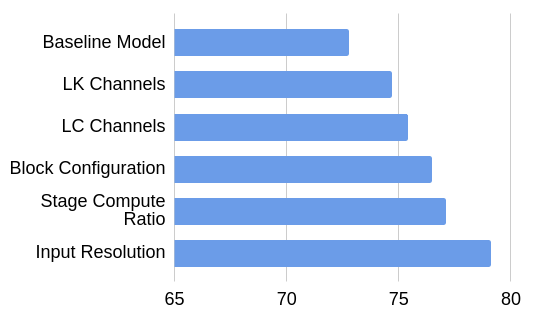}
  }
  \vspace{-0.1cm}
  \caption{Progressive specialization of network architecture for 3D object detection in Autonomous vehicle.}
  \label{fig:progress}
\end{figure}

\begin{table}[t!]
\centering
\vspace{0.3cm}
\caption{Comparison of Scaled Model Configurations}
\label{tab:config}
\resizebox{\columnwidth}{!}{%
\begin{tabular}{|l|c|c|c|}
\hline
\multicolumn{1}{|c|}{\textbf{\begin{tabular}[c]{@{}c@{}}Model \\ Variant\end{tabular}}} &
  \textbf{\begin{tabular}[c]{@{}c@{}}Number of \\ blocks \\ per stage\end{tabular}} &
  \textbf{\begin{tabular}[c]{@{}c@{}}Number of channels\\  in large kernel\\  convolution\end{tabular}} &
  \textbf{\begin{tabular}[c]{@{}c@{}}Number of channels\\ in large channel\\  convolution\end{tabular}} \\ \hline
Tiny  & 1, 5, 2, 1   & 64                 & 64, 128, 256, 128  \\ \hline
Small & 1, 7, 4, 1   & 64                 & 64, 128, 256, 128  \\ \hline
Base  & 2, 14, 8, 2  & 128                & 128, 256, 512, 256 \\ \hline
Large & 4, 28, 16, 4 & 128, 128, 256, 128 & 128, 320, 512, 256 \\ \hline
\end{tabular}%
}
\vspace{0.3cm}
\end{table}

Our design decisions have collectively enhanced the model's performance, resulting in an improvement over the baseline model from $72.8\%$ mAP to an optimal model with $79.2\%$ mAP. Fig. \ref{fig:progress} summarizes improvement in model performance with each stage of tuning. 

\subsection{Experimenting with Architectural Scaling}
We construct different variants of the model, Tiny/Small/Base/Large (denoted at T/S/B/L for simplicity), to be of similar complexities to CoAtNet \cite{coatnet}. The Tiny \& Small variants are suitable for online processing whereas the Base \& Large variants are targeted for offline processing. Our base model is the end product of the progressive architecture refinement. The variants only differ in the number of channels, and the number of blocks in each stage. We summarize the configurations in Table \ref{tab:config}.

As seen from our results in Fig. \ref{fig:main_result} the optimized model scales well across the spectrum, offering versatility for different online and offline processing needs.

\section{Discussion}

Adapting the model to AV data led to significant design modifications. In this section we summarize the key findings.
\begin{itemize}
\item Prioritizing feature extraction and amplification in the initial stages with more number of blocks/channels, where larger feature map facilitates long range detection, proved to be more beneficial than the traditional approach of expanding the network width in deeper layers.

\item Reducing the model capacity in later stages reduces the complexity while maintaining accuracy, a further advantage facilitated by the smaller number of classes compared to models trained on large class datasets.

\item We underscore the importance of tailoring stage-compute-ratio to individual tasks and datasets. This optimal configuration not only delivers superior performance for our specific objective but also opens doors for potential application to other tasks involving small object detection or limited class sets.

\item By integrating a self-attention module into our CNN architecture, we achieve additional performance gains, demonstrating the strong potential of hybrid approaches.

\item Interestingly, no single model achieves peak performance across all object classes. This observation highlights the potential for further exploration and targeted model specialization to achieve best class-wise accuracy.

\item Operating the network at higher resolution contributed to improved accuracy, despite the quadrupled computational overhead. While we simply remove the initial downsample layer of the  model, there are multiple ways to effectively leverage the high resolution details which could likely lead to better accuracy and lower computational overhead.

\end{itemize}

\section{Conclusions}
\label{sect:conclusions}
This work presents a series of design changes that adapts the model to address the needs of AV dataset. Our results show that the customization has led to significant accuracy improvement over the baseline, culminating in an optimized family of model architectures. Our findings not only pave the way for further advancements in AV perception but also highlight the value of customization to achieve best performance on any new dataset.

%% file: main_result.tex
\begin{figure*}[!ht]
  \centering
  \begin{subfigure}{.5\textwidth}
  \centering
    \includegraphics[width=.9\linewidth]{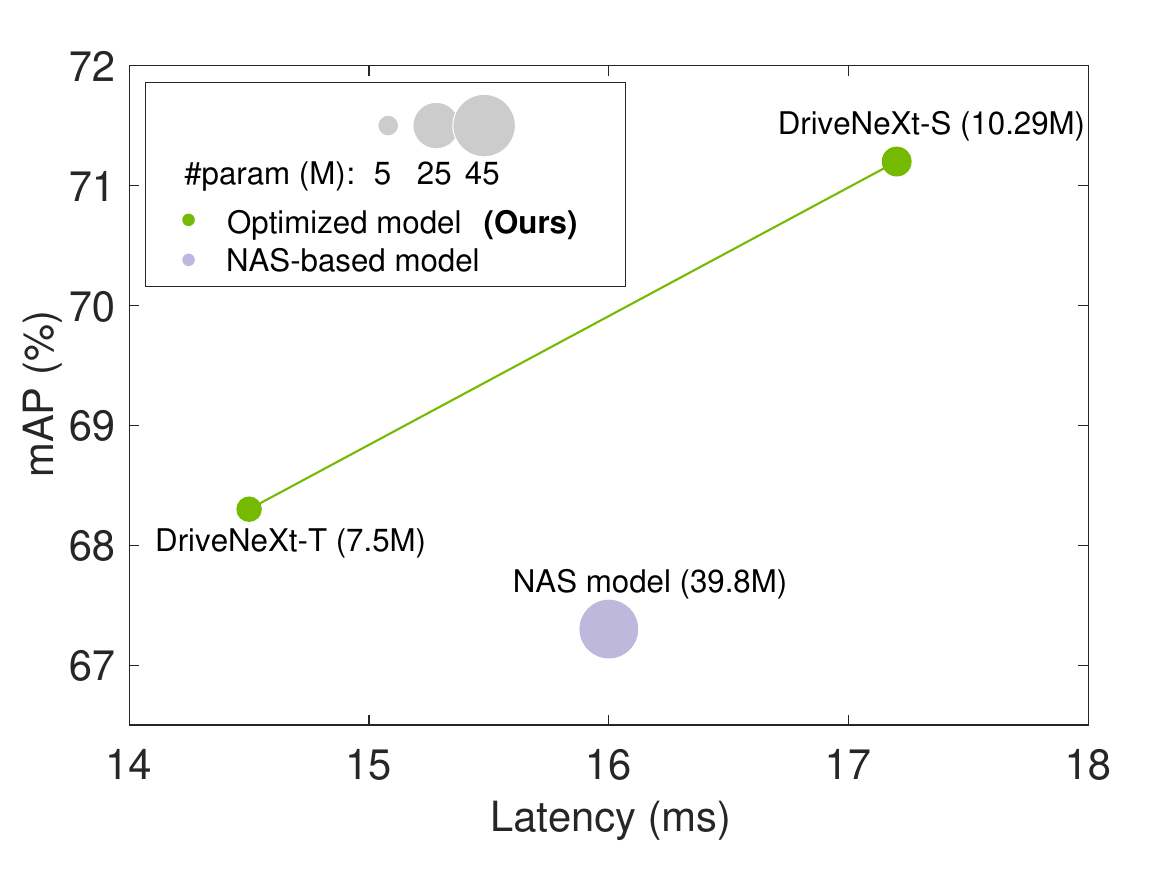}
    \caption{}
  \end{subfigure}%
  \begin{subfigure}{.5\textwidth}
  \centering
    \includegraphics[width=.9\linewidth]{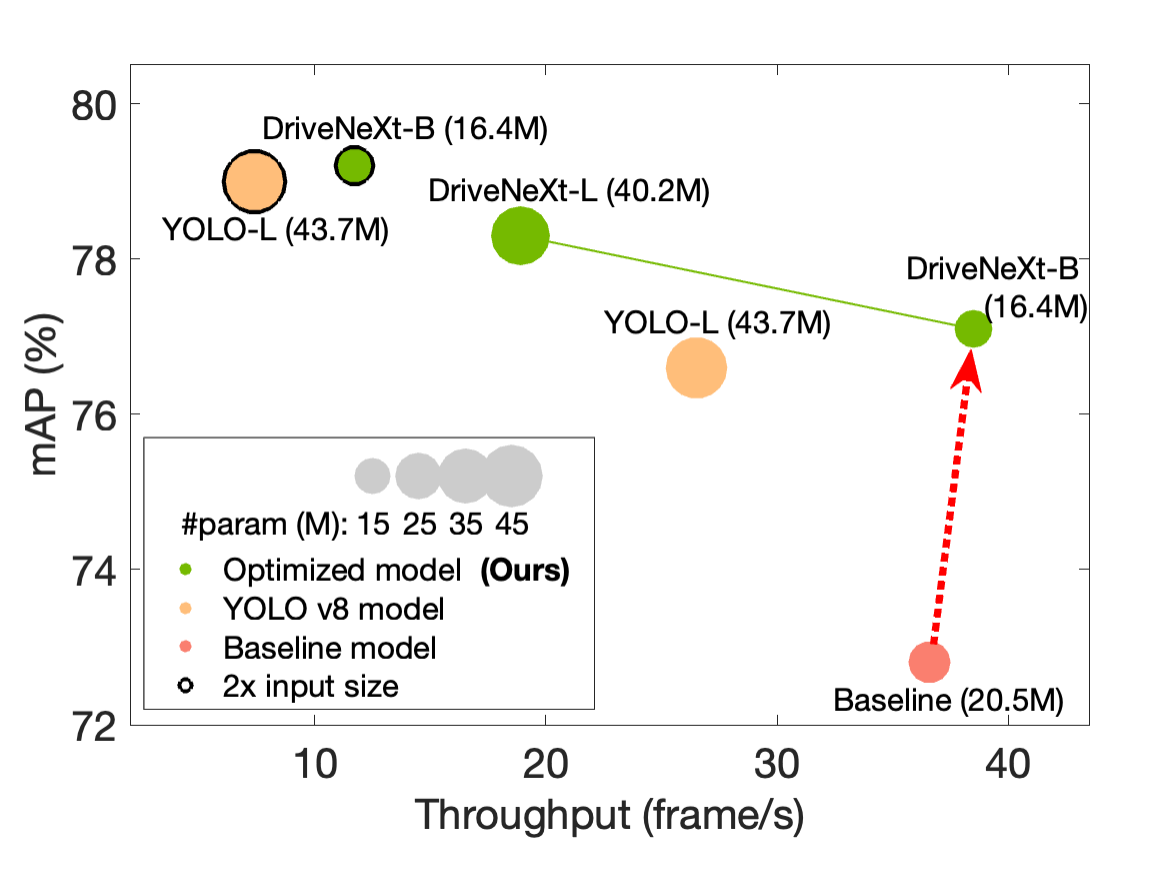}
    \caption{}
  \end{subfigure}
  \vspace{-0.5cm}
  \caption{Our models adapted for AV needs exhibit superior performance both in terms of accuracy and throughput compared to the un-optimized baseline. We provide performance of NAS based architecture and off-the-shelf YOLO v8 architecture for reference. (a) Online models profiled on AGX Orin GPU with batch size $1$. (b) Offline models profiled on Ampere A100 GPU with batch size $32$.
  }
  \label{fig:main_result}
\end{figure*}